\documentclass[letter, 10 pt, conference]{ieeeconf}  % Comment this line out if you need a4paper

\usepackage[pdftex]{graphicx}
\usepackage{amsmath}
\usepackage{amsfonts}
\usepackage{subfigure}
\usepackage{tabularx}
\usepackage{cite}
\usepackage[bookmarks=false,colorlinks,linkcolor=black,citecolor=black,urlcolor=black]{hyperref}

\graphicspath{{Graphics/}}

\IEEEoverridecommandlockouts                              % This command is only needed if
\begin{document}

\title{\LARGE \bf
Real-Time Force Control of an SEA-based Body Weight Support Unit with the 2-DOF Control Structure
\author{Yubo Sun$^{1,2}$, Yuqi Lei$^{1,2,3}$, Wulin Zou$^{1,2}$, Jianmin Li$^{4}$, Ningbo Yu$^{1,2 *}$ } 
\thanks{This work is supported by the National Natural Science Foundation of China (61720106012, 61403215) and the Fundamental Research Funds for the Central Universities.}%
\thanks{$^1$ Institute of Robotics and Automatic Information Systems, Nankai University, Haihe Education Park, Tianjin 300353, China.}
\thanks{$^2$ Tianjin Key Laboratory of Intelligent Robotics, Nankai University, Haihe Education Park, Tianjin 300353, China.}%
\thanks{$^3$ Department of Electrical and Computer Engineering, Carnegie Mellon University, 5000 Forbes Ave, Pittsburgh, PA 15213, USA.}%
\thanks{$^4$ School of Mechanical Engineering, Tianjin University, Tianjin 300072, China.}%
\thanks{$^*$ Corresponding author. Email: nyu@nankai.edu.cn, Phone: +86 (0)22 2350 3960 ext. 801.}%
}

\maketitle
\thispagestyle{empty}
\pagestyle{empty}

%%%%%%%%%%%%%%%%%%%%%%%%%%%%%%%%%%%%%%%%%%%%%%%%%%%%%%%%%%%%%%%%%%%%%%%%%%%%%%%%

\begin{abstract}
	
Body weight support (BWS) is a fundamental technique in  rehabilitation. 
Along with the dramatic progressing of rehabilitation science
and engineering, BWS is quickly evolving with new initiatives and has attracted
deep research effort in recent years. We have built up a novel gravity
offloading system, in which the patient is allowed to move freely in the three-dimensional Cartesian space and receives support against gravity. Thus, the
patients, especially for those that suffer from neurological injury such as
stroke or spinal cord injury, can focus their residual motor control
capabilities on essential therapeutic trainings of balance and gait. The
real-time force control performance is critical for the BWS unit to provide
suitable support and avoid disturbance. In this work, we have re-designed our
BWS unit with a series elastic actuation structure to improve the human-robot interaction
performance. Further, the 2 degrees of freedom (2-DOF) control approach was
taken for accurate and robust BWS force control. Both simulation and
experimental results have validated the efficacy of the BWS design and
real-time control methods.

\end{abstract}

\section{Introduction}

Along with the dramatic progressing of rehabilitation science and engineering, body weight support (BWS) is attracting novel initiative and deep research effort~\cite{Courtine_2017, Awai_2017}. 

BWS has been a fundamental technique in rehabilitation practice for a long time. The primary function of the BWS unit is to offload partial body weight from the patients with lower limb dysfunction, especially those that suffer from neurological injury such as stroke or spinal cord injury, so that they can focus their residual motor control capabilities on therapeutic trainings of balance and gait function. Body weight support has been an essential component in many rehabilitation robotic systems~\cite{develop1,develop2,ZeroG,FLOAT,lokomat}.

A key performance index for BWS is the force control accuracy, but it has not been well addressed in conventional BWS systems. The series elastic actuation (SEA) structure~\cite{Pratt1995}, which transforms force control into displacement control, makes it possible to realize high quality gravity offloading without the need of a force sensor. This provides a more convenient way for accurate BWS force control, and we have realized a body weight support system with cable-driven SEA in our previous work~\cite{Yang2016ARM, yu_2017_AAS_AGOS, AGOS1}.

Cable actuation has the advantages of flexible installation, high force-to-weight ratio, low inertia, etc~\cite{SEA_adopt2,Yu2017AA,Yu2017JAS}, and has been used for many robotic applications, such as actuation in the environment of magnetic resonance scanner~\cite{cable1}, lower extremity powered exoskeleton~\cite{Veneman2006}, etc. The control of cable actuation for physical human-robot interaction has also been studied in~\cite{SEA_adopt2,cable2,cable_SEA1}. 

However, we are confronted with control challenges with our cable-driven SEA based body weight support system. Conventional PID method did not suffice our requirement for accurate and fast force control when the position is altered by the subject during walking~\cite{AGOS1}. Accurate modeling of the system brought in nonlinearities, and the resulted nonlinear controllers are heavily model-dependent, making it difficult to achieve robustness~\cite{yu_2017_AAS_AGOS}. 

The 2-DOF control method presented by Horowitz and then further developed by Qiu, Zhou and Huang~\cite{2dof_method1,2dof_method2}, is a promising solution. With the 2-DOF structure, the controller can be designed in such a way that the two problems of fast and accurate reference tracking as well as system robustness can be simultaneously handled.

The paper is organized as follows. Section II describes the hardware platform and modeling of the designed cable-driven SEA. Details of the controller design are given in Section III. Simulations, practical experiments and results are presented in Section IV. Finally, section V concludes the paper.

\section{Hardware Platform and Modeling}

\subsection{Mechanical Design}

\begin{figure}[!thbp]
	\centering
	\includegraphics[width=0.8\columnwidth]{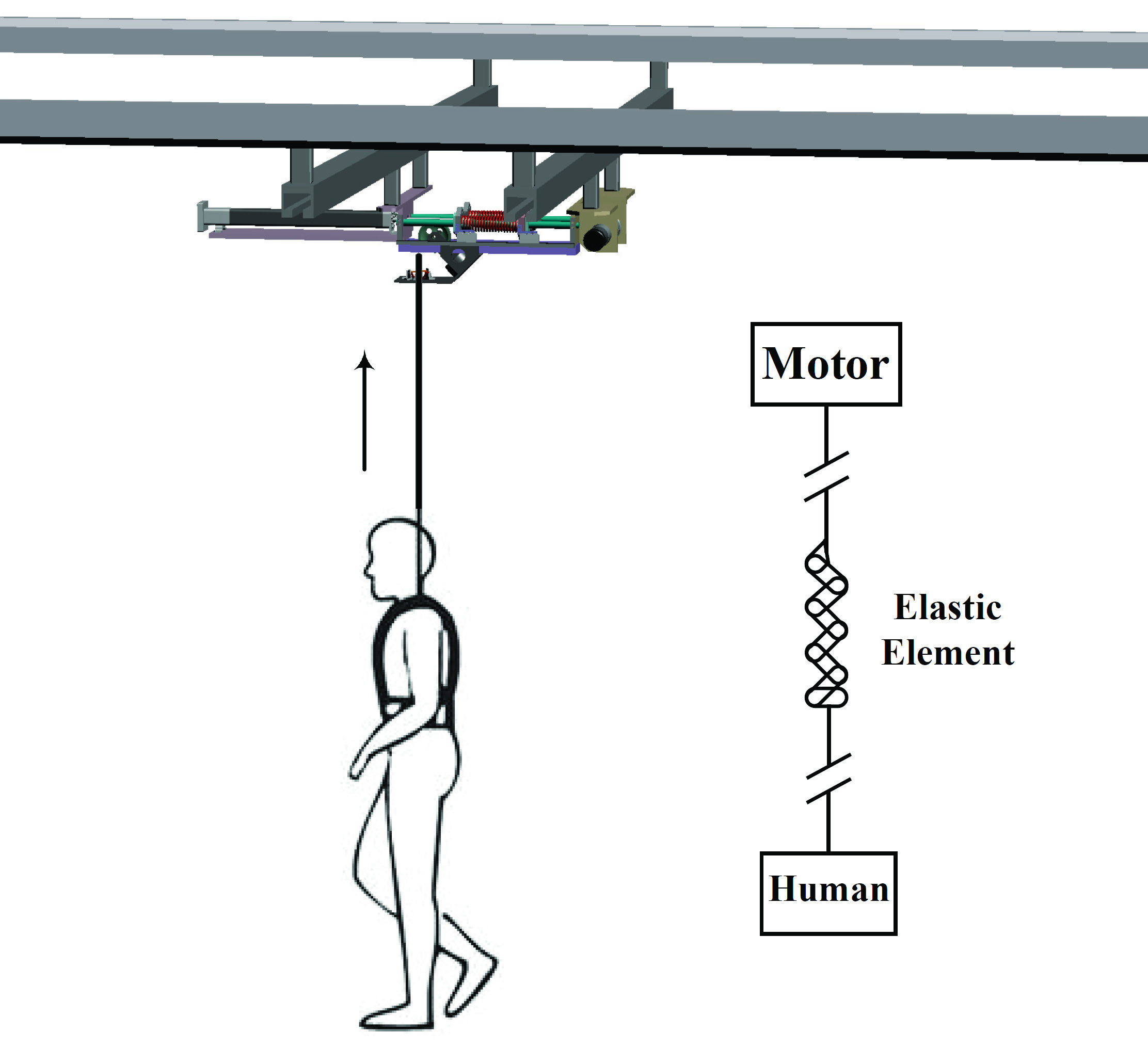}
	\caption{The schematic diagram of the body weight support system based on series elastic actuation.}
	\label{fig_BWS_with_human}
\end{figure}

\begin{figure}[!htbp]
	\centering
	\includegraphics[width=0.8\columnwidth]{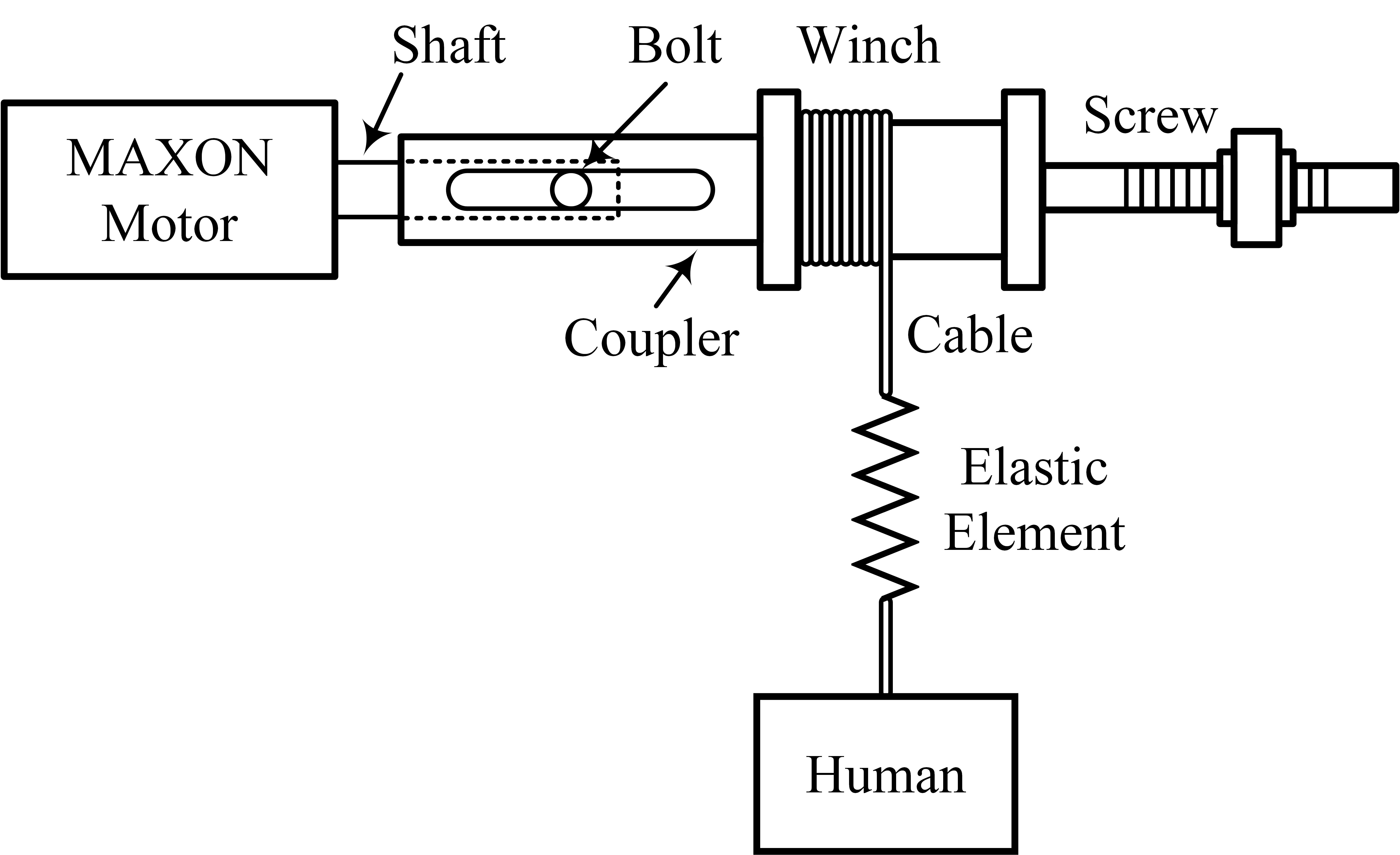}
	\caption{Mechanical design of the SEA unit.}
	\label{fig_design_winch}
\end{figure}

The designs of the body-weight support training system and the cable-driven SEA are illustrated in Fig.~\ref{fig_BWS_with_human} and Fig.~\ref{fig_design_winch}. The patient is allowed to move freely in the three-dimensional Cartesian space and receives support against gravity. In addition, this system is also eligible to provide forward force if desired by the training therapy~\cite{Courtine_2017,yu_2017_AAS_AGOS}.

For actuation of body weight support, a DC brushless rotary motor (Maxon EC-i 40 100W 48V) was used as the velocity source. The motor is equipped with a 3-channel incremental encoder (HEDL 5540 500CPT). A planetary gear head (GP42C 3.5:1) was installed on its output shaft. The motor velocity was managed by a servo controller (Accelnet Micro Panel).

A coupler was mounted at the end of the motor shaft. When the motor rotates, a winch at the other end of the coupler will rotate together and move in the horizontal direction so that the cable is pulled to provide the tension force. A displacement sensor (WPS-500-MK30-E) together with a spring were installed to measure the elastic deformation and the tension force, which can be calculated with the spring stiffness and the deformation.

\subsection{Real-Time Data Acquisition and Control}

\begin{figure}[!htbp]
	\centering
	\includegraphics[width=0.95\columnwidth]{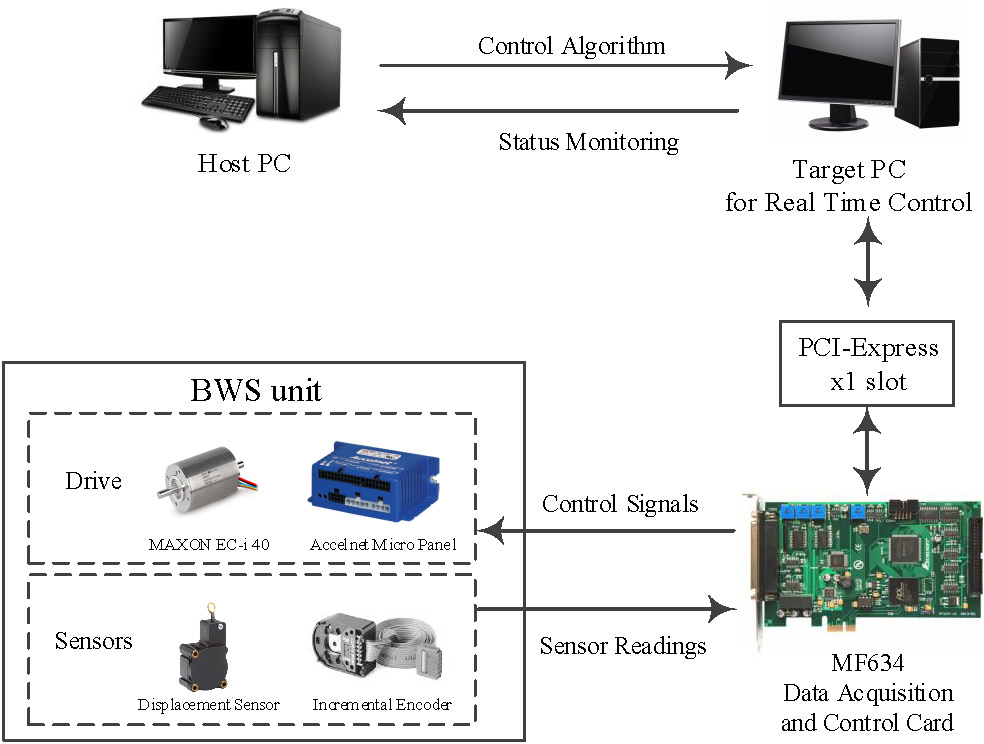}
	\caption{The real-time data acquisition and control framework.}
	\label{fig_Real_Time_Control}
\end{figure}

The MATLAB/Simulink Real-Time toolbox was used for real-time data acquisition and control of the system. The control algorithm was firstly designed in MATLAB/Simulink, compiled in the host PC, and then downloaded to the target PC to run in real-time. A data acquisition board (HUMUSOFT MF634 PCI-Express multifunction I/O card) was inserted into the PCI-Express slot of the target PC, reading in sensor measurements and sending analog control signals to the motor servo controller. This process is shown in Fig.~\ref {fig_Real_Time_Control}.

\subsection{System Modeling}

The transfer function $G(s)$ from the control signal ($u$, v) to the motor velocity ($\omega_m$, rad/s) is identified as
\begin{equation}
{G(s)} = \frac{\omega_m(s)}{u(s)} = \frac{69788}{s^2+39.2s+840}.
\label{equ_g1s1}
\end{equation}

Then, the transfer function $P(s)$ from control signal $u$ to cable tension force $F$ is
\begin{equation}
\begin{aligned}
{P(s)} &= \frac{F(s)}{u(s)} = {G(s)} \times \frac{1}{s}  \times \frac{r}{K_g} \times K_s\\
& = \frac{183440}{s^3+39.2s^2+840s}.
\end{aligned}
\label{equ_ps}
\end{equation}

Here, ${K_g = 3.5}$ is the reduction ratio of the planetary gear, ${K_s = 460 \text{ N/m}}$ is the spring constant. ${r=0.02 \text{ m}}$ is the radius of the cable winch.

\section{Controller Design}

\begin{figure}[!htbp]
	\centering
	\includegraphics[width=1\columnwidth]{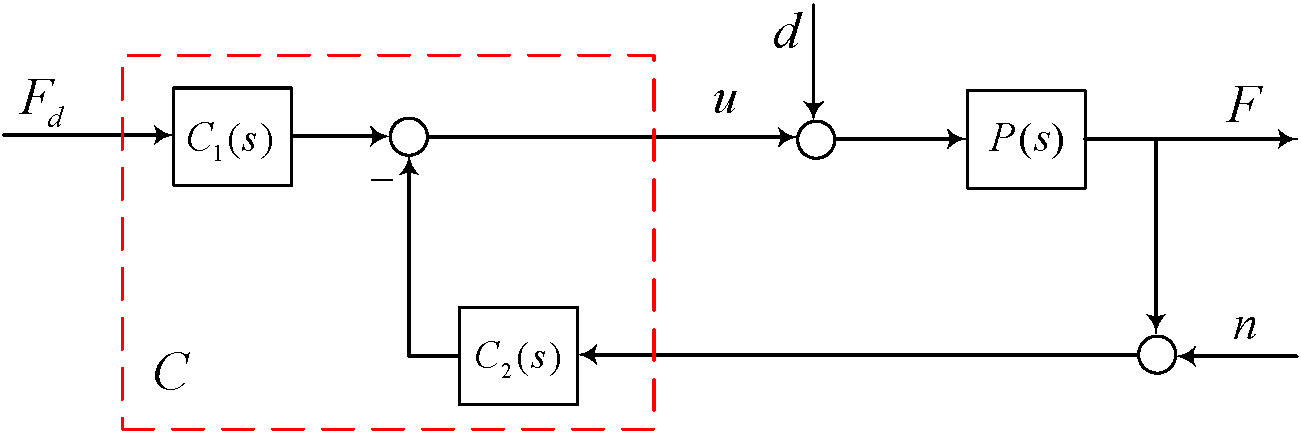}
	\caption{Diagram of the 2-DOF control structure.}
	\label{fig_2DOF_REAL}
\end{figure}

The proposed force control method with the 2-DOF structure is depicted in Fig.~\ref{fig_2DOF_REAL}. $F_d$ and $F$ are the desired and actual output force respectively. $u$ denotes the control signal of the motor. $d$ and $n$ are the disturbance and noise signal respectively. $C = \left[C_1(s),~C_2(s)\right]$ represents the torque controller.

Systematic design procedures for stabilizing 2-DOF controller are given as the following steps~\cite{2dof_method2}:
\begin{description}
  \item[Step 1]: $P(s)$ is denoted as
  \[P(s) = \frac{{b(s)}}{{a(s)}} = \frac{{{b_1}{s^{n - 1}} +  \cdots  + {b_n}}}{{{a_0}{s^n} + {a_1}{s^{n - 1}} +  \cdots  + {a_n}}}.\]
  \item[Step 2]: Factorize the following polynomial to obtain a stable $d_\rho(s)$.
  \[{\rho ^2}a( - s)a(s) + b( - s)b(s) = {d_\rho }( - s){d_\rho }(s)\]
  where $\rho>0$ is a relative weight to $u(t)$ and tracking error $e(t)$.
  \item[Step 3]: Factorize the following polynomial to obtain a stable $d_{\lambda ,k}(s)$.
  \[-{k^2}s^2a( - s)a(s) + {\lambda ^2}b( - s)b(s) = {d_{\lambda ,k}}( - s){d_{\lambda ,k}}(s)\]
  where $\lambda>0$ is a relative weight to $d(t)$ and $r(t)$, while $k>0$ is a relative weight to $n(t)$ and $r(t)$.
  \item[Step 4]: Strictly proper subcontroller ${C_2}(s) = {q(s)}/{p(s)}$ is a type 1 controller with order $n+1$, and can be obtained by spectral factorization such that
      \[a(s)p(s) + b(s)q(s) = {d_\rho }(s){d_{\lambda ,k}}(s).\]
  \item[Step 5]: Subcontroller $C_1(s)$ is obtained as
  \[{C_1}(s) = \frac{{{d_\rho }(0)}}{{b(0)}}\frac{{{d_{\lambda ,k}}(s)}}{{p(s)}}.\]

\end{description}

The three parameters $\rho$, $\lambda$ and $k$ tune the performance for tracking, disturbance and noise rejection, and energy consumption.

\section{Results}

The 2-DOF controller has been validated with both simulation and experiments, in comparison with the typical PID controller.
We set parameters of the 2-DOF controller according to the system performance requirements. Parameters of the PID controller were tuned by MATLAB toolbox.

\subsection{Simulation and Results}

\begin{figure}[!htbp]
	\centering
	\includegraphics[width=0.9\columnwidth]{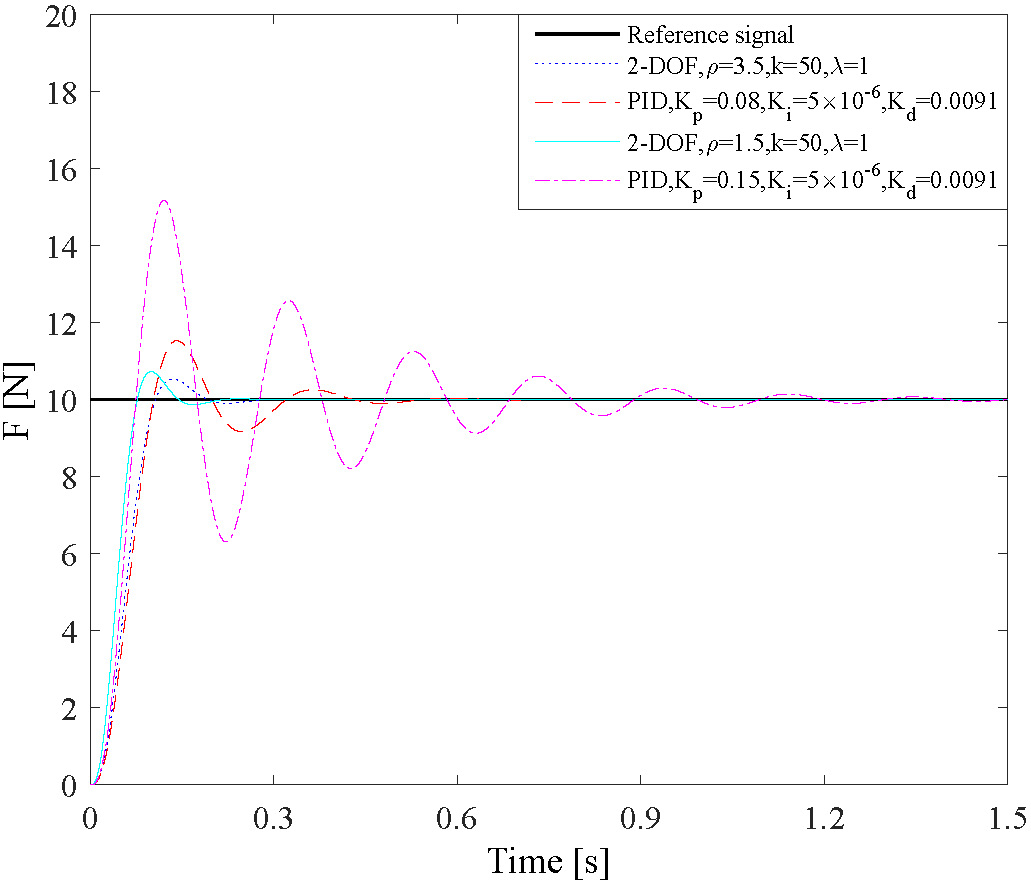}
	\caption{Responses of the two methods in the first simulation.}
	\label{fig_sim_step1}
\end{figure}

In the first simulation, two groups simulations have been conducted to compare the performance of the PID and 2-DOF control method. Parameters of the two controllers are shown in Table~\ref{parameters of controllers}, where $K_p,~K_i,~K_d$ are the parameters of the PID controller. A 10 N step signal was set as the reference input. The results are shown in Fig.~\ref{fig_sim_step1}. In each group, the rising time of PID control and 2-DOF control were close, while other performance indexes clearly showed that the 2-DOF controller excelled the PID controller. Besides, quantified performance metrics are summarized in Table~\ref{quantified_behaviour}, which also revealed better performance of the 2-DOF method. 

\begin{table}[!htbp]
	\centering
	\caption{Controller Parameters for the Two Methods in the First Simulation.}
	\label{parameters of controllers}
	{\renewcommand{\arraystretch}{1.5} \setlength\tabcolsep{3.0mm}
		\begin{tabular}{cc|cc}
			\hline  \hline
			\multicolumn{2}{c|}{Group A}  & \multicolumn{2}{c}{ Group B}  \\
			\hline
			2-DOF & PID & 2-DOF & PID \\
			\hline 
			$\rho$=0.091 & ${K_{p}}$=0.08 & $\rho$=0.067 & ${K_{p}}$=0.15\\
			\hline
			$\kappa$=0.1715 & ${K_{i}}$=${5\times10^{-6}}$ & $\kappa$=0.1322 & ${K_{i}}$=${5\times10^{-6}}$\\
			\hline
			$\lambda$=0.53 & ${K_{d}}$=${9.1\times10^{-3}}$ & $\lambda$=0.726 & ${K_{d}}$=${9.1\times10^{-3}}$\\
			\hline \hline
		\end{tabular}}
\end{table}
	
\begin{table}[!htbp]
  \centering
  \caption{Quantified Performance Metrics of the Two Methods in the First Simulation.}
  \label{quantified_behaviour}
  {\renewcommand{\arraystretch}{1.5} \setlength\tabcolsep{2.5mm}
  \begin{tabular}{c|cc|cc}
  \hline  \hline
   & \multicolumn{2}{c|}{Group A}  & \multicolumn{2}{c}{ Group B}  \\
   \hline
   & 2-DOF & PID &  2-DOF &  PID \\
  \hline 
  Rising Time (s) & 0.091 & 0.093 & 0.067 & 0.070\\
  \hline
  Settling Time (s) & 0.1715 & 0.3885 & 0.1322 & 1.0478\\
  \hline
  Overshoot (N) & 0.530 & 1.530 & 0.726 & 5.164\\
  \hline
  Overshoot Ratio ($\%$) & 5.30 & 15.30 & 7.26 & 51.64\\
  \hline
  Steady State Force (N) & 10 & 10 & 10 & 10 \\
  \hline
  Steady State Error (N) & 0 & 0 & 0 & 0\\
  \hline \hline
  \end{tabular}}
\end{table}

\begin{figure}[!htbp]
	\centering
	\includegraphics[width=0.9\columnwidth]{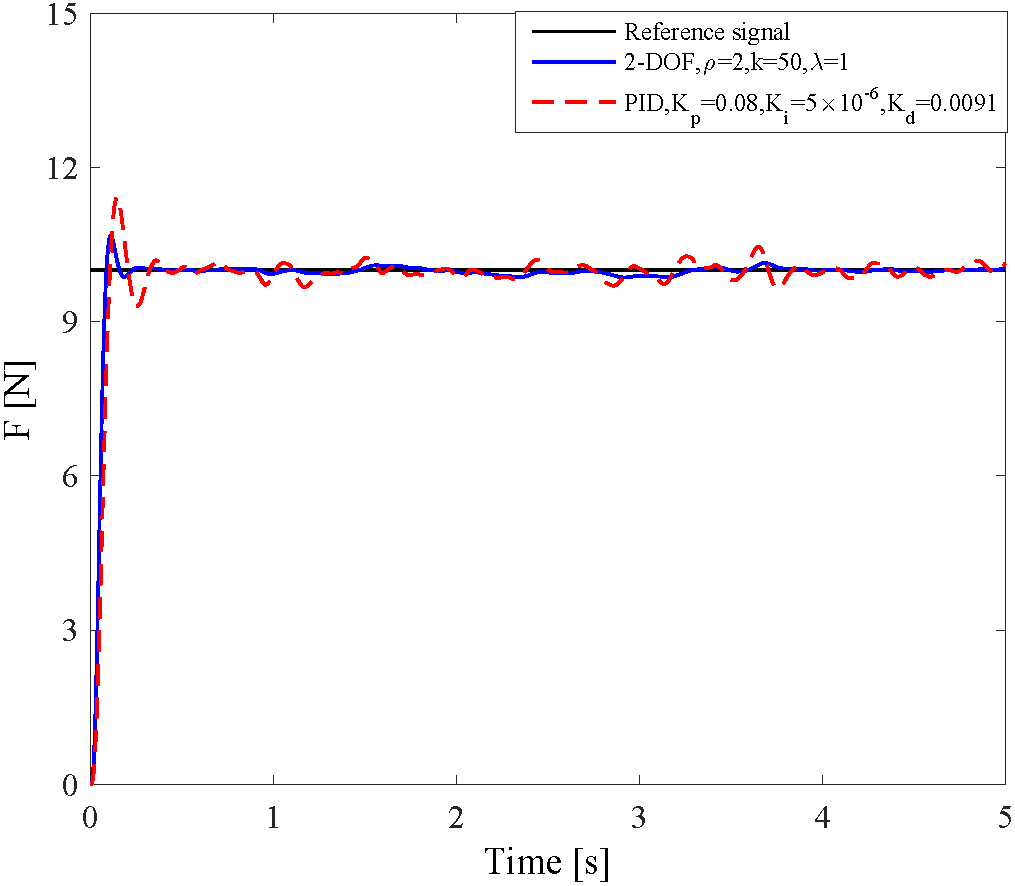}
	\caption{Responses of the two methods in the second simulation.}
	\label{fig_sim_step2}
\end{figure}

\begin{figure}[!htbp]
\centering
		\includegraphics[width=0.48\columnwidth]{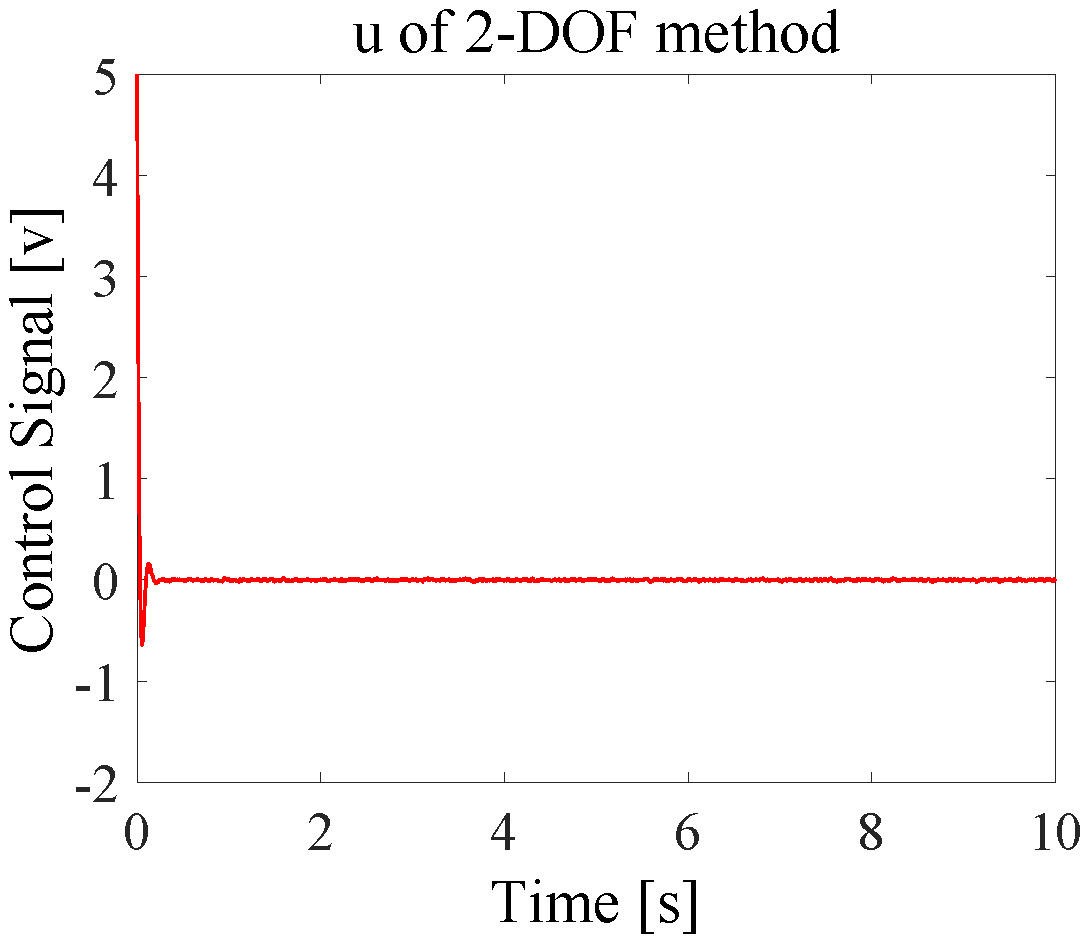}~
		\includegraphics[width=0.48\columnwidth]{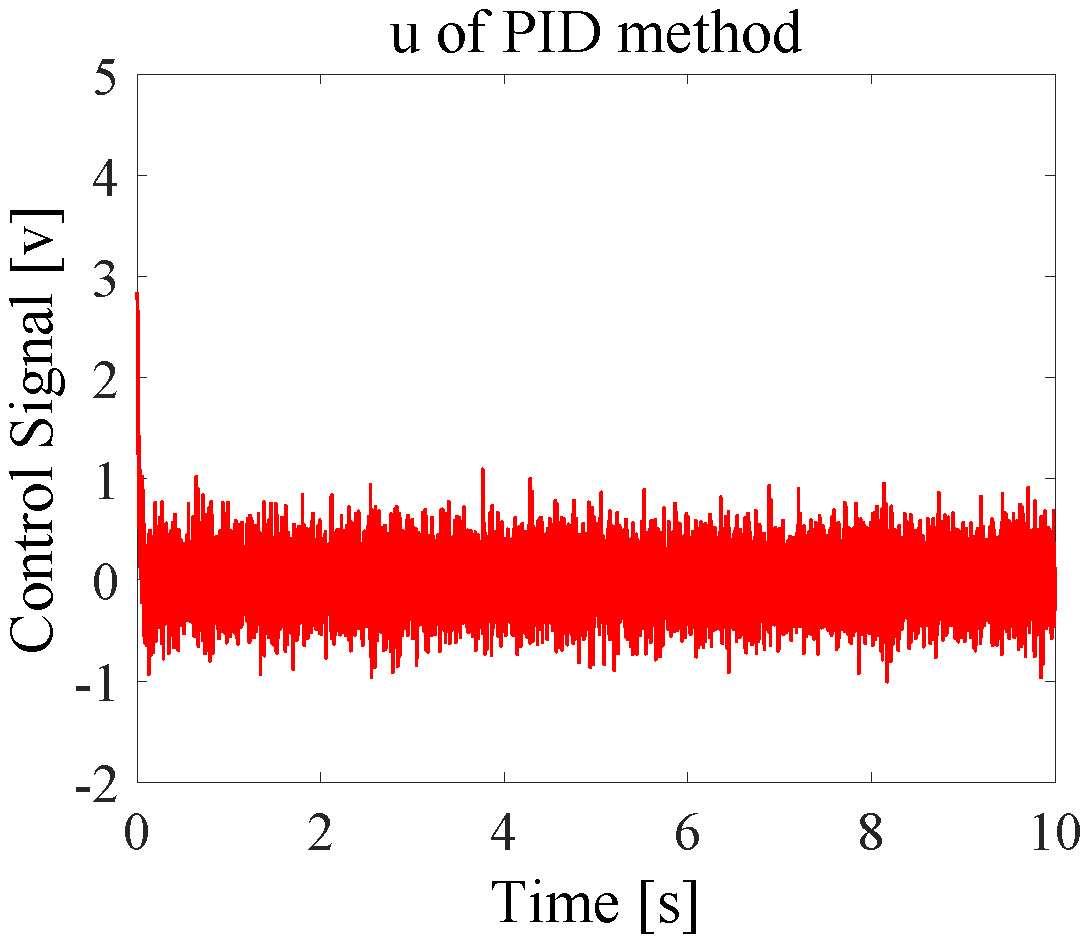}
	\caption{Control signals of the two methods in the second simulation.}
	\label{fig_sim_step_u}
\end{figure}

The second simulation was performed with disturbance and noise to further illustrate the superiority of the 2-DOF controller. In this simulation, the disturbance ${d}$ and noise ${n}$ were Gaussian white noise. As shown in Fig.~\ref{fig_sim_step2}, it is obvious that the overshoot of the PID method is much larger. In Fig.~\ref{fig_sim_step_u}, the 2-DOF controller is much better for disturbance rejection.

\begin{figure}[!htbp]
	\centering
	\includegraphics[width=0.9\columnwidth]{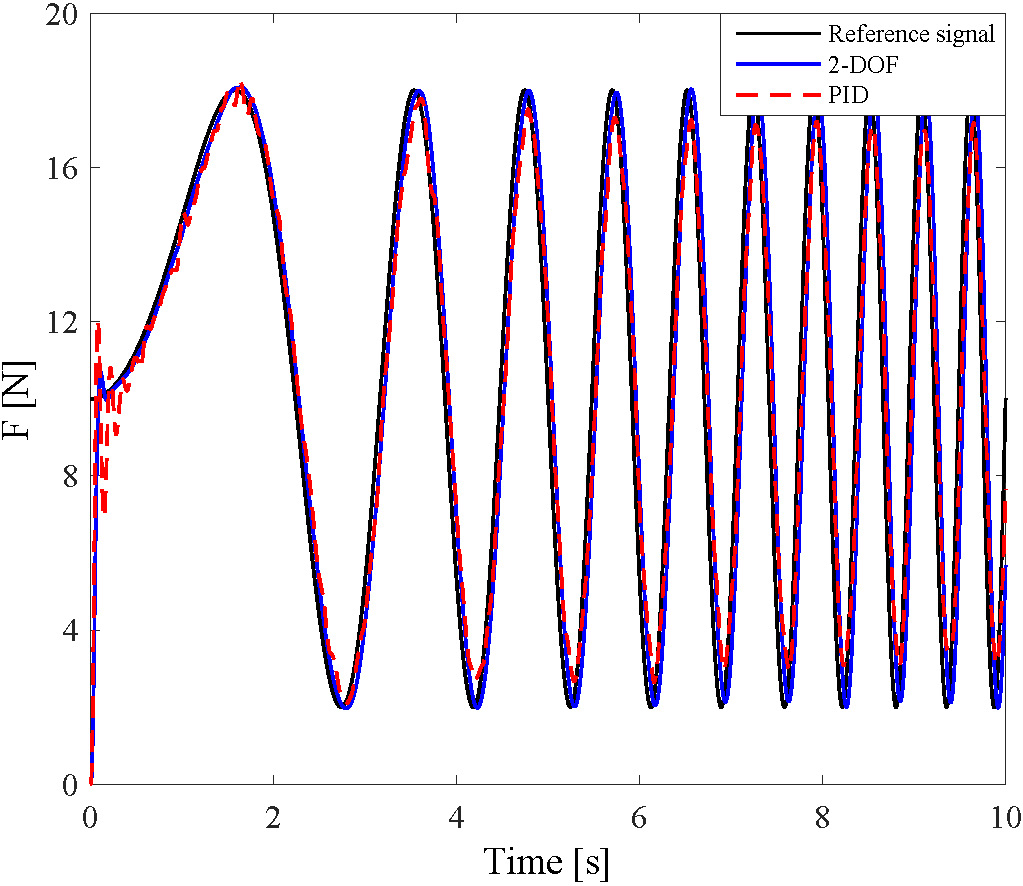}
	\caption{Chirp responses of the two methods in the third simulation.}
	\label{fig_sim_chirp}
\end{figure}

\begin{figure}[!htbp]
		\centering
		\includegraphics[width=0.48\columnwidth]{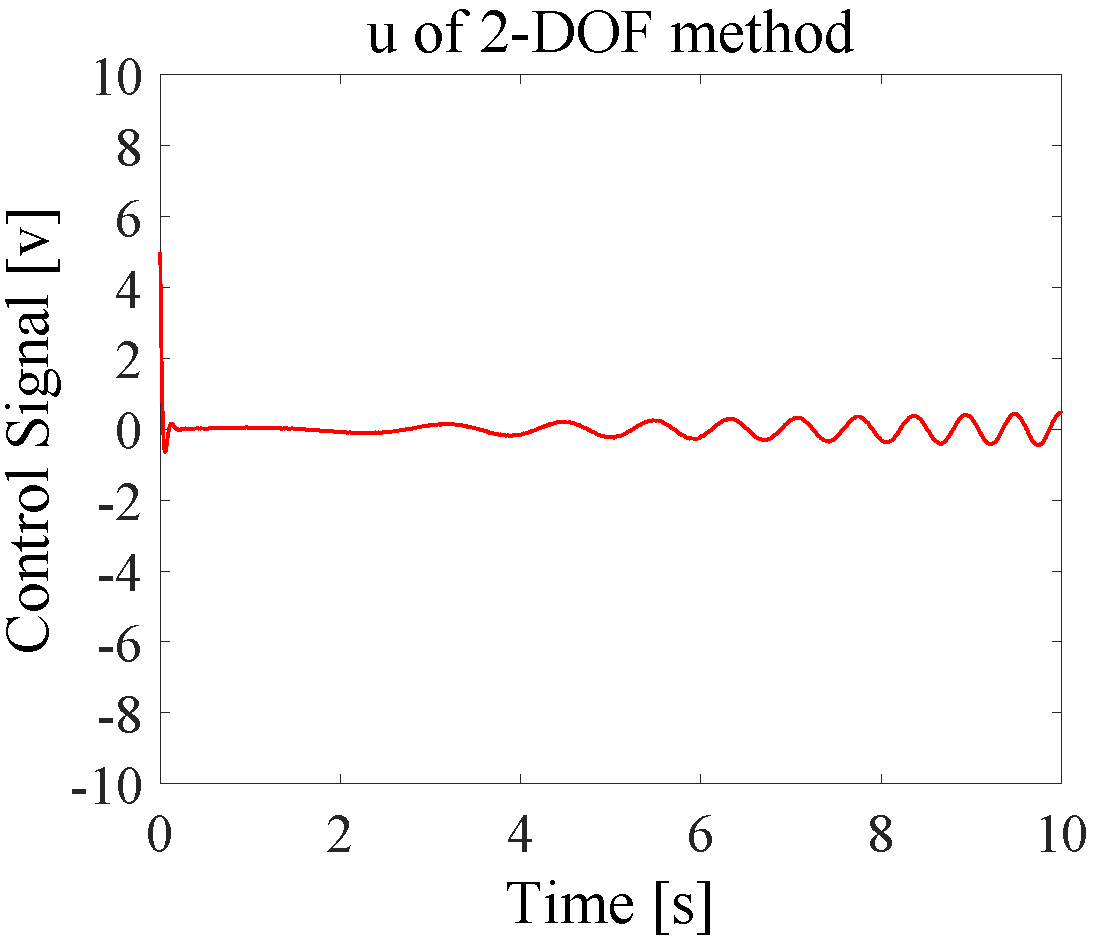}
		\includegraphics[width=0.48\columnwidth]{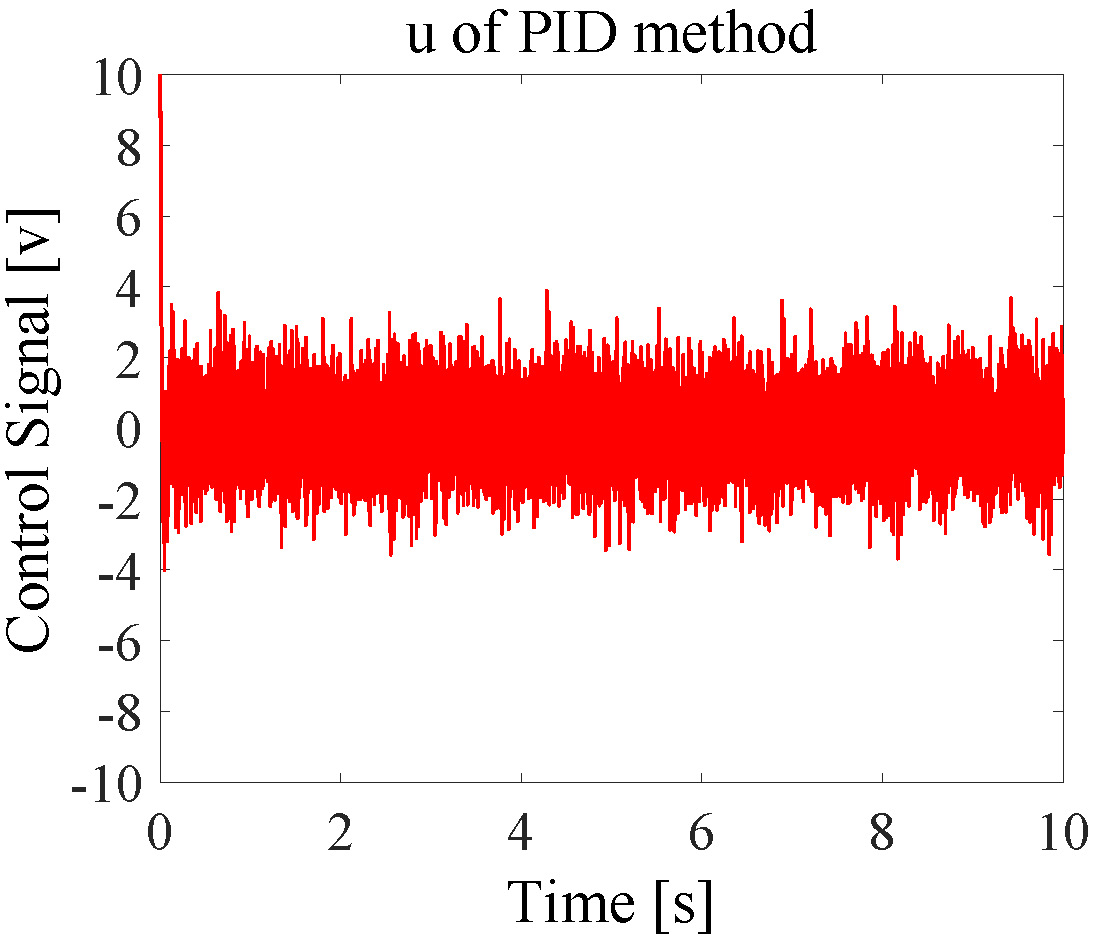}
	\caption{Control signals of the two methods in the third simulation.}
	\label{fig_sim_chirp_u}
\end{figure}

In the last simulation, a linear chirp signal was used as the reference input to evaluate tracking performance at different frequencies. This linear chirp signal's frequencies ranged from 0 to 2 Hz in 10 seconds. From Fig.~\ref{fig_sim_chirp}, the tracking performance of the 2-DOF controller was better than the PID controller. From Fig.~\ref{fig_sim_chirp_u}, the robustness of the 2-DOF controller was better than the PID controller, and less energy was consumed.

\subsection{Experiments and Results}

In the experiments, the parameters of 2-DOF and PID controllers for experiments are shown in Table~\ref{parameters_controllers_exp}. The output side of the SEA is fixed to the ground rather than a human subject. Besides the disturbance and noise inherently from the electro-mechanical system, we also purposely added disturbance ${d}$ and noise ${n}$, both were Gaussian signals.

\begin{table}[!htbp]
	\centering
	\caption{Controller Parameters for the Experiments.}
	\label{parameters_controllers_exp}
	{\renewcommand{\arraystretch}{1.5} \setlength\tabcolsep{3.0mm}
		\begin{tabular}{cccc}
			\hline\hline	 	
			2-DOF & $\rho$=3 & $\kappa$=2 & $\lambda$=10  \\
			\hline
			PID & ${K_{p}}$=0.08 & ${K_{i}}$=${5\times10^{-6}}$ & ${K_{d}}$=${9.1\times10^{-3}}$ \\
			  
			\hline\hline
		\end{tabular}}
	\end{table}

\begin{figure}[!htbp]
	\centering
	\includegraphics[width=0.9\columnwidth]{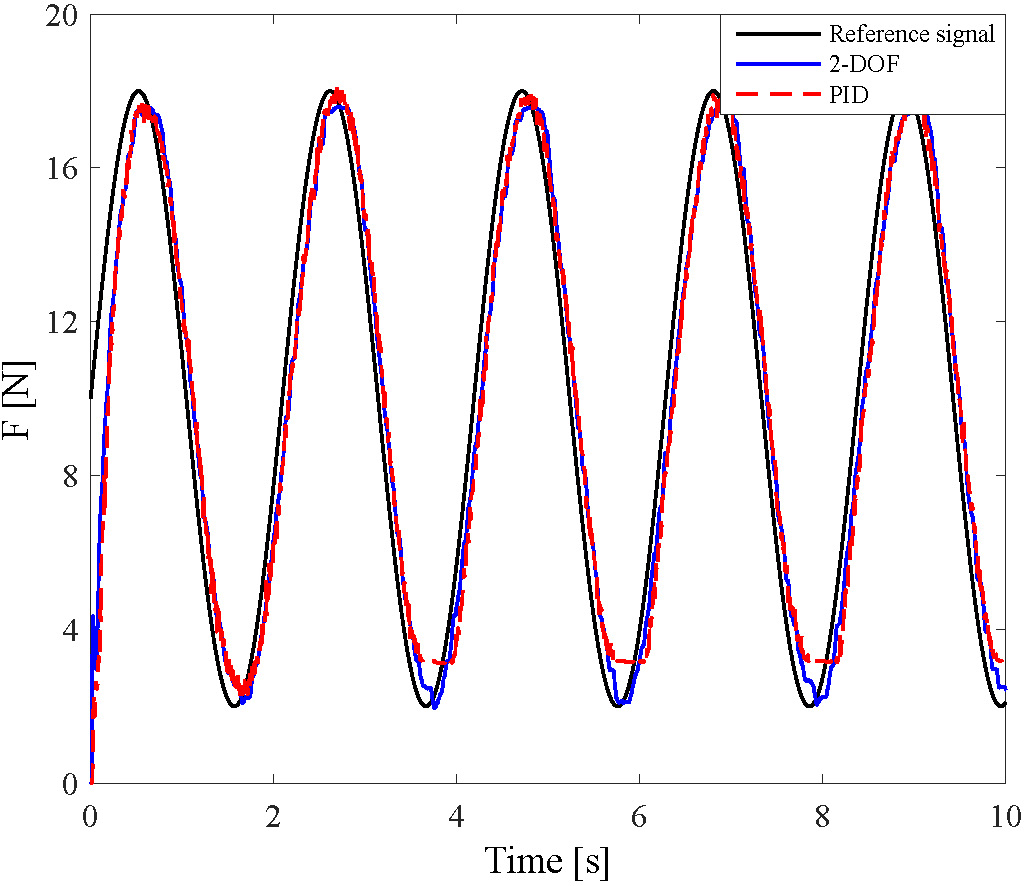}
	\caption{Sinusoidal responses of the two methods in the first experiment.}
	\label{fig_exp_sine}
\end{figure}

\begin{figure}[!htbp]
		\centering
		\includegraphics[width=0.48\columnwidth]{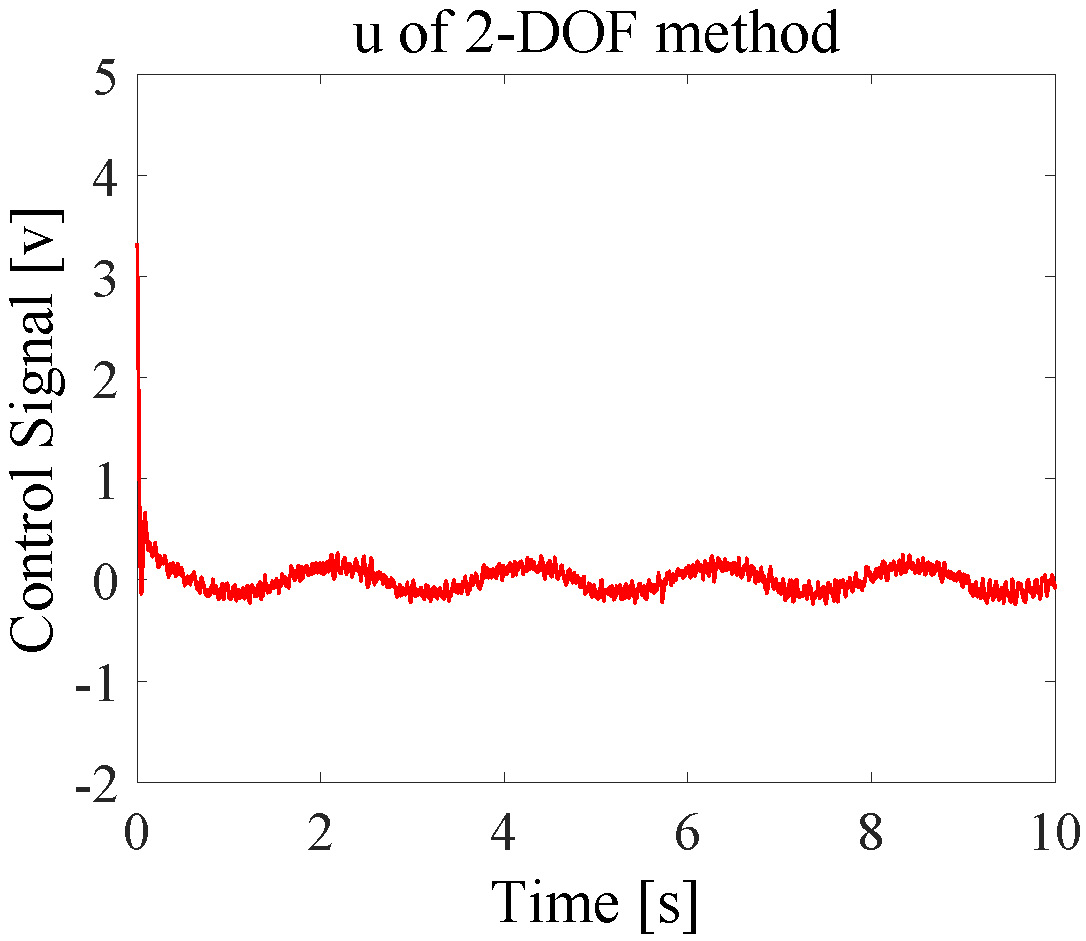}~
		\includegraphics[width=0.48\columnwidth]{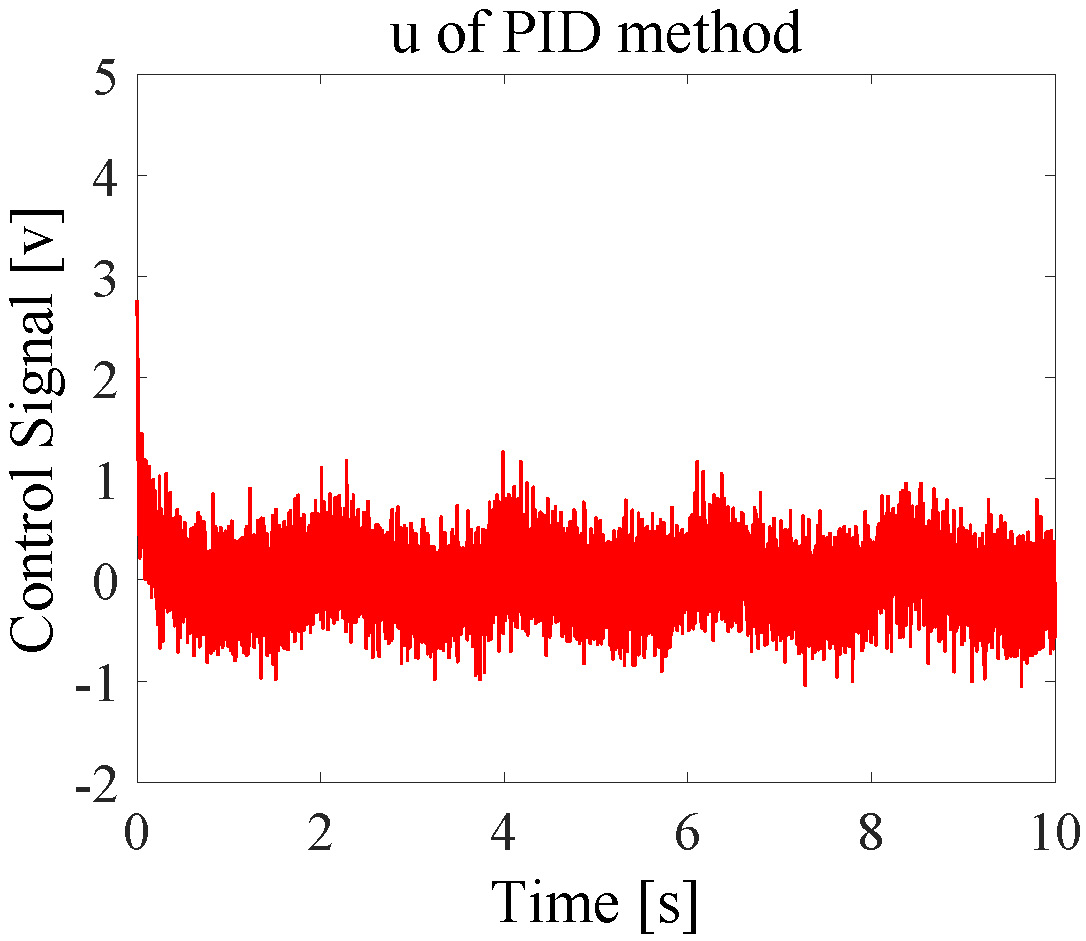}
	\caption{Control signals of the two methods in the first experiment.}
	\label{fig_exp_sine_u}
\end{figure}

During the first experiment, the reference input signal was a sinusoidal signal with frequency of 0.5 Hz. The PID controller produced poor performance as shown in Fig.~\ref{fig_exp_sine}. The 2-DOF controller had much better tracking performance. In Fig.~\ref{fig_exp_sine_u}, the robustness of the 2-DOF controller was better than the PID controller and less energy was consumed. These were consistent with the simulation in the previous section.

\begin{figure}[!htbp]
	\centering
	\includegraphics[width=0.9\columnwidth]{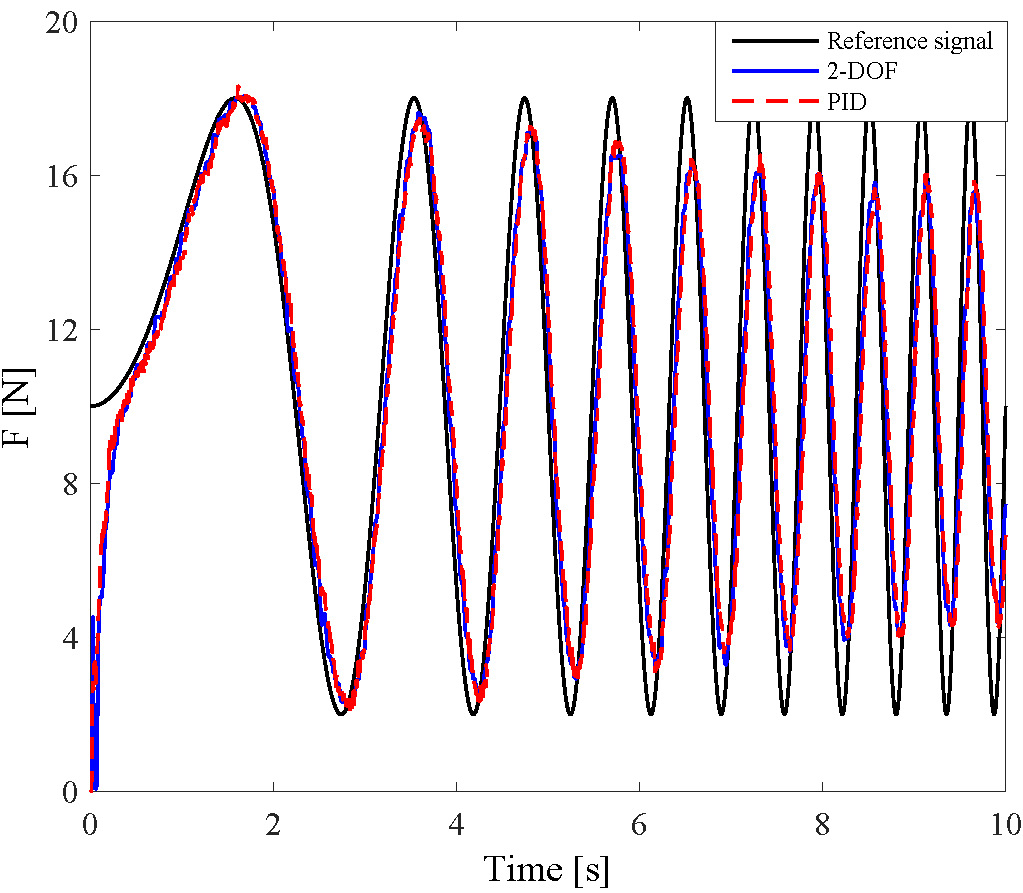}
	\caption{Chirp responses of the two methods in the second experiment.}
	\label{fig_exp_chirp}
\end{figure}

\begin{figure}[!htbp]
		\centering
		\includegraphics[width=0.48\columnwidth]{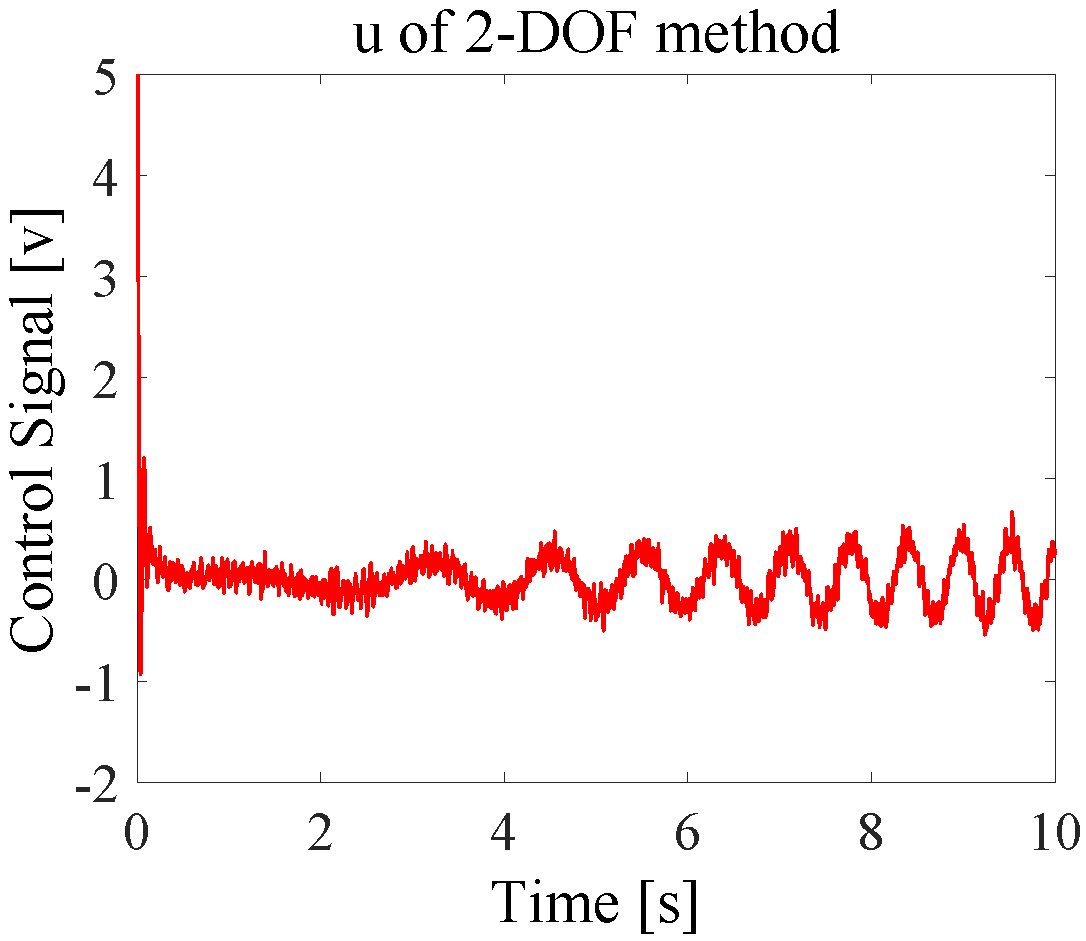}~
		\includegraphics[width=0.48\columnwidth]{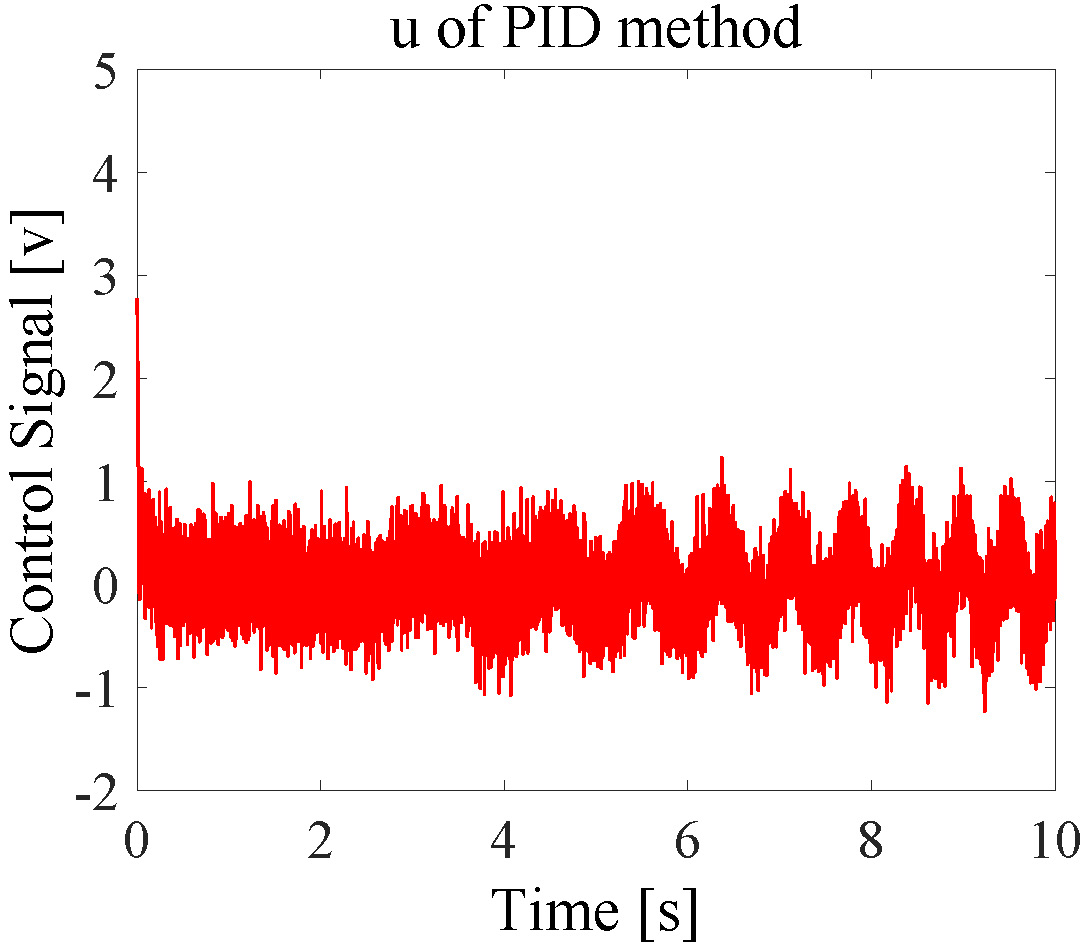}
	\caption{Control signals of the two methods in the second experiment.}
	\label{fig_exp_chirp_u}
\end{figure}

During the last experiment, a linear chirp signal, which was the same as the one in the simulation, was used as the reference input. As the experimental results shown in Fig.~\ref{fig_exp_chirp}, the tracking performance of the PID and 2-DOF controllers were close. However, as shown in Fig.~\ref{fig_exp_chirp_u}, the 2-DOF controller brought better system robustness, and less energy was consumed.  

\section{Conclusion}

The body weight support unit is a key component in many rehabilitation systems with the function of offloading the patient by a controlled amount of force against gravity. Cable-driven series elastic actuation has been designed to realize accurate control of the cable tension force. Further, the 2-DOF control structure has been taken to achieve good tracking performance and disturbance/noise rejection simultaneously, with guaranteed robustness. Both simulation and experimental results validated the efficacy of the mechanical design and real-time control methods.

\bibliography{References} 

\end{document}